\documentclass[letterpaper, 10pt, conference]{ieeeconf}  

\IEEEoverridecommandlockouts                              
\usepackage{graphicx} 
\usepackage{color}
\usepackage{tikz}
\usepackage{cite}
\usepackage{subcaption}
\usepackage{multirow}
\usepackage{tabularx}
\usepackage{epsfig}
\usepackage{mathptmx} 
\usepackage{amsmath} 
\usepackage{amssymb}  
\usepackage{soul}
\usepackage{dblfloatfix}
\graphicspath{{./figs/}}

\title{\LARGE \bf Effects of Different Hand-Grounding Locations on Haptic Performance with a Wearable Kinesthetic Haptic Device}

 \author{Sajid Nisar$^{1,2}$, Melisa Orta Martinez$^{1}$, Takahiro Endo$^{2}$, Fumitoshi Matsuno$^{2}$, and Allison M. Okamura$^{1}$ 
 \thanks{This work was supported in part by National Science Foundation grant 1812966.}
 \thanks{$^{1}$
 	S.~Nisar, M.~Orta Martinez and A.~M.~Okamura are with the Department of Mechanical Engineering, Stanford University,
 	Stanford, CA 92305, USA
     	}%
 \thanks{$^{2}$
 	S.~Nisar, T.~Endo and F.~Matsuno are with the Department of Mechanical Engineering \& Science, Kyoto University, Kyoto 615-8200, Japan
         {\tt\small nisar.sajid.78v$@$kyoto-u.jp}
			}%
 }

\begin{document}

\maketitle
\thispagestyle{empty}
\pagestyle{empty}

\begin{abstract}
	Grounding of kinesthetic feedback against a user's hand can increase the portability and wearability of a haptic device. However, the effects of different hand-grounding locations on haptic perception of a user are unknown. In this paper, we investigate the effects of three different hand-grounding locations -- back of the hand, proximal phalanx of the index finger, and middle phalanx of the index finger -- on haptic perception using a newly designed wearable haptic device. The novel device can provide kinesthetic feedback to the user's index finger in two directions: along the finger-axis and in the finger's flexion-extension movement direction. We measure users' haptic perception for each grounding location through a psychophysical experiment for each of the two feedback directions. Results show that among the studied locations, grounding at proximal phalanx has a smaller average Just Noticeable Difference for both feedback directions, indicating more sensitive haptic perception. The realism of the haptic feedback, based on user ratings, was highest with grounding at the middle phalanx for feedback along the finger axis, and at the proximal phalanx for feedback in the flexion-extension direction. Users identified the haptic feedback as most comfortable with grounding at the back of the hand for feedback along the finger axis and at the proximal phalanx for feedback in the flexion-extension direction. These findings show that the choice of grounding location has significant impact on the user's haptic perception and qualitative experience. The results provide insights for designing next-generation wearable hand-grounded kinesthetic devices to achieve better haptic performance and user experience in virtual reality and teleoperated robotic applications.   
\end{abstract}



\section{INTRODUCTION}
	\begin{tikzpicture}[overlay]
	\node [right, text width= 7.3in, align=left] at (-.5,15.7) {Preprints of the \textbf{IEEE Robotics and Automation Letters (RAL)} paper presented at the \\\textbf{2019 International Conference on Robotics and Automation (ICRA)}, Palais des congres de Montreal, Montreal, Canada,\\May 20-24, 2019. The final version of the article can be accessed at DOI: 10.1109/LRA.2018.2890198
	};
	\end{tikzpicture}The majority of the existing haptic devices providing kinesthetic feedback are world grounded~\cite{pacchierotti2017wearable}. They offer numerous advantages like high forces and torques, many degrees of freedom (DoF), and a wide dynamic range. These features allow such devices to provide more realistic haptic renderings compared to tactile haptic devices that only stimulate the skin. 
	However, the world-grounded kinesthetic haptic devices generally have a large footprint as well as limited portability and wearability, which limits their application and effectiveness for many virtual and real-world applications. World-grounded haptic devices also offer a limited range of motion to the user due to the scaling of weight and friction with increased size~\cite{sucho@2016}.
	
	On the other hand, wearable haptic devices must be portable and typically offer a large range of motion. But, majority of the existing wearable haptic devices are tactile in nature and provide feedback in the form of vibration or skin deformation. They are commonly grounded against the user's fingertip or the nearby region~\cite{pacchierotti2017wearable}. Though tactile feedback is capable of providing directional cues and aiding users in completing various tasks, it may not be sufficient to perform certain tasks, such as the suture knot-tying in robot-assisted surgery~\cite{okamura2009haptic}, and manipulating objects in virtual reality~\cite{burdea1999keynote}. As demonstrated by Suchoski et al.~\cite{sucho@2016} in their study, kinesthetic feedback is capable to give more sensitive haptic information to carry out a grasp-and-lift task than the skin deformation feedback (a form of tactile feedback). Similarly, the role of kinesthetic (force) feedback in surgical training and skill development looks very promising~\cite{okamura2009haptic}.


    Kinesthetic haptic devices, that are not world grounded but instead impart feedback by grounding forces against the user's hand (\emph{hand-grounded haptic devices}), provide a solution to challenges of portability, wearability and limited workspace in kinesthetic haptic devices. As noted by Pacchierotti et al.~\cite{pacchierotti2017wearable}, the primary advantage of wearable kinesthetic devices is their small form factor as compared to the world-grounded devices.  Similarly, body-grounded kinesthetic devices, i.e. Exoskeletons, could be another potential solution, but they generally encumber the user movement and are difficult to don and doff. 
    However, designing these hand or body-grounded devices is challenging due to the need for increased forces/torques and number of degrees-of-freedom (DoF) in comparison to the fingertip tactile devices. Additionally, the effects of hand-grounded kinesthetic feedback on users' perception and haptic experience are still unknown.
	
	
	There exist numerous examples of hand-grounded kinesthetic haptic devices, including~\cite{jadhav2017soft,fontana2009mechanical,springer2002design,leonardis2015emg,nycz2016design,allotta2015development,ma2015rml,kim2016hapthimble,fu2011design,lambercy2013design,stergiopoulos2003design,lelieveld2006design,cempini2015powered,agarwal2015index,aiple2013pushing,tanaka2002wearable,polygerinos2015soft,stetten2011hand,bouzit2002rutgers,choi2018claw,choi2016wolverine,khurshid2014wearable}. These devices are either grounded against the back of the hand~\cite{jadhav2017soft,fontana2009mechanical,springer2002design,leonardis2015emg,nycz2016design,allotta2015development,ma2015rml,kim2016hapthimble,fu2011design,lambercy2013design,stergiopoulos2003design,lelieveld2006design,cempini2015powered,agarwal2015index,aiple2013pushing}, act like a glove ~\cite{tanaka2002wearable,polygerinos2015soft,stetten2011hand}, are grounded against the user's palm~\cite{bouzit2002rutgers,choi2018claw}, or are grounded against the user's fingers~\cite{choi2016wolverine,khurshid2014wearable}. To the best of our knowledge, there exists no device that can be grounded against different locations on the user's hand or a study that explains the effect of different grounding locations on the user's haptic perception and qualitative experience with kinesthetic (force) feedback.

	
	\begin{figure}
		\fontfamily{cmss}\selectfont
		\centering
		\def\svgwidth{.9\columnwidth}
		{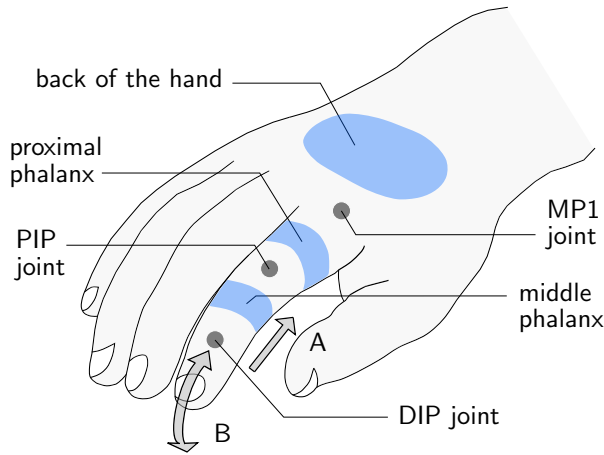}
		\caption{Three potential grounding locations on the user's hand: Back of the hand, Proximal Phalanx, and Middle Phalanx of the index finger. Arrows indicate directions of applied kinesthetic feedback on the fingertip: (A) along the finger axis and (B) in flexion-extension.}
		\label{fig:hand}
	\end{figure}
	
	\begin{figure*}[ht]
		\fontfamily{cmss}\selectfont
		\centering
		\begin{subfigure}{0.32\textwidth}
			\def\svgwidth{1\textwidth}
			{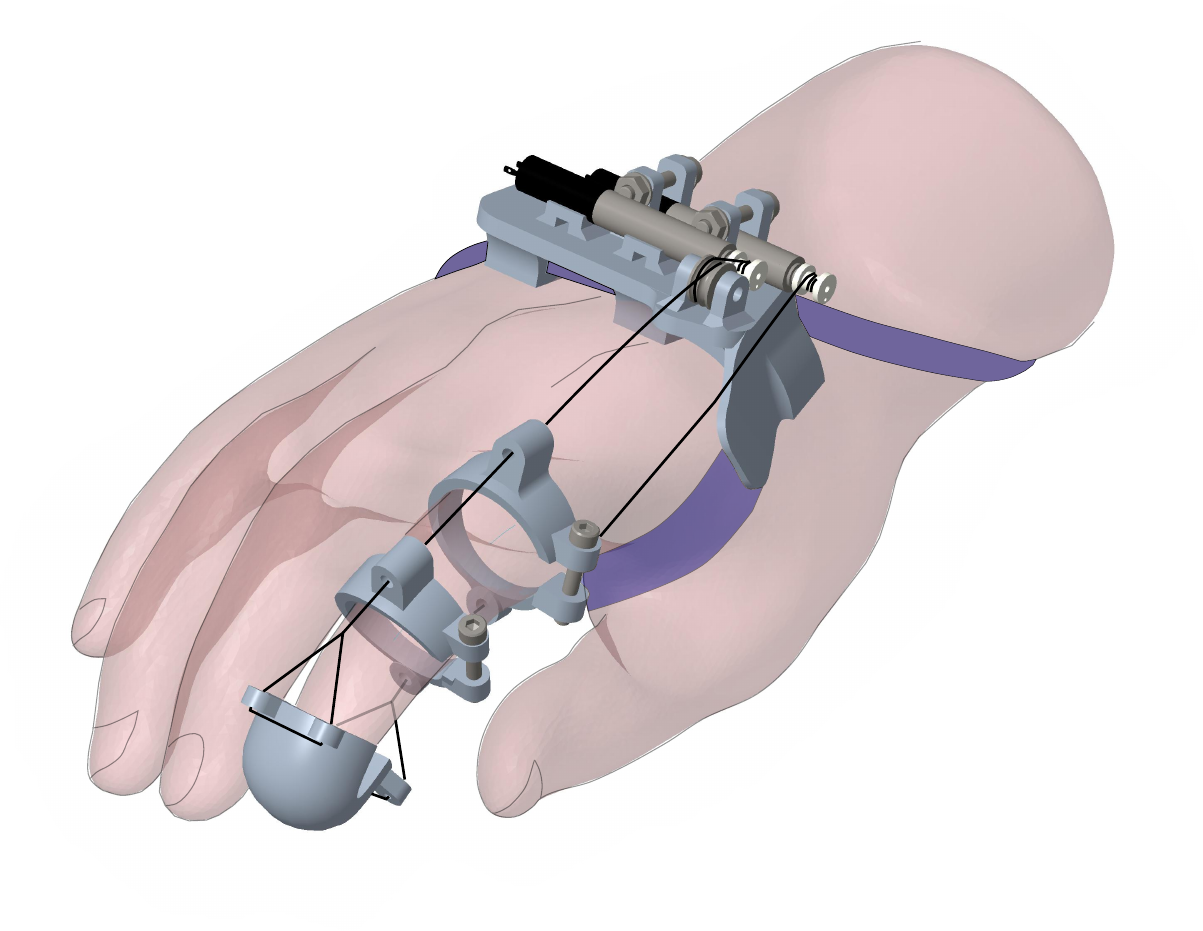}
			\vspace{-.6cm}
			\caption{}
			\label{fig:a}
		\end{subfigure}
		\centering
		\begin{subfigure}{0.32\textwidth}
			\def\svgwidth{1\columnwidth}
			{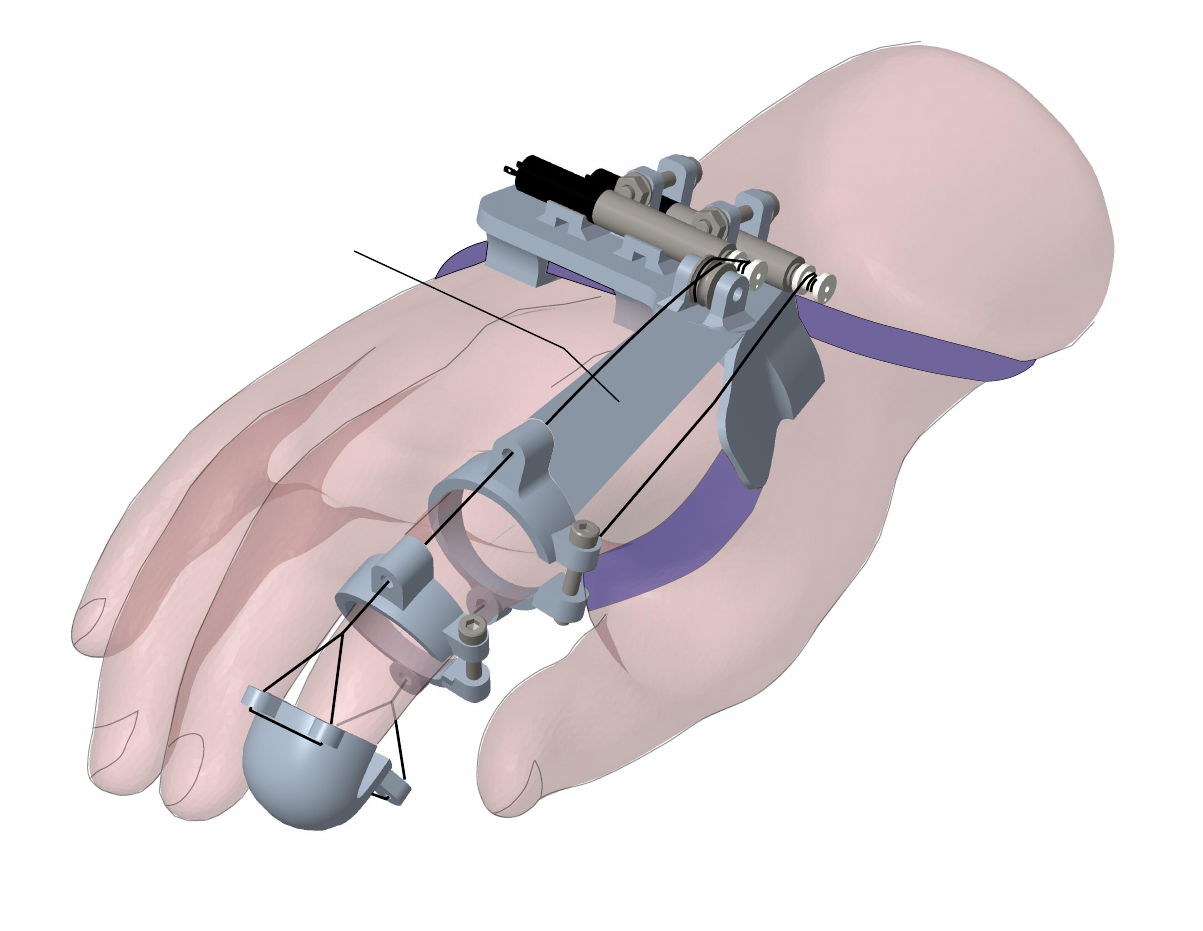}
			\caption{}
			\label{fig:b}
		\end{subfigure}
		\begin{subfigure}{0.32\textwidth}
			\def\svgwidth{1\columnwidth}
			{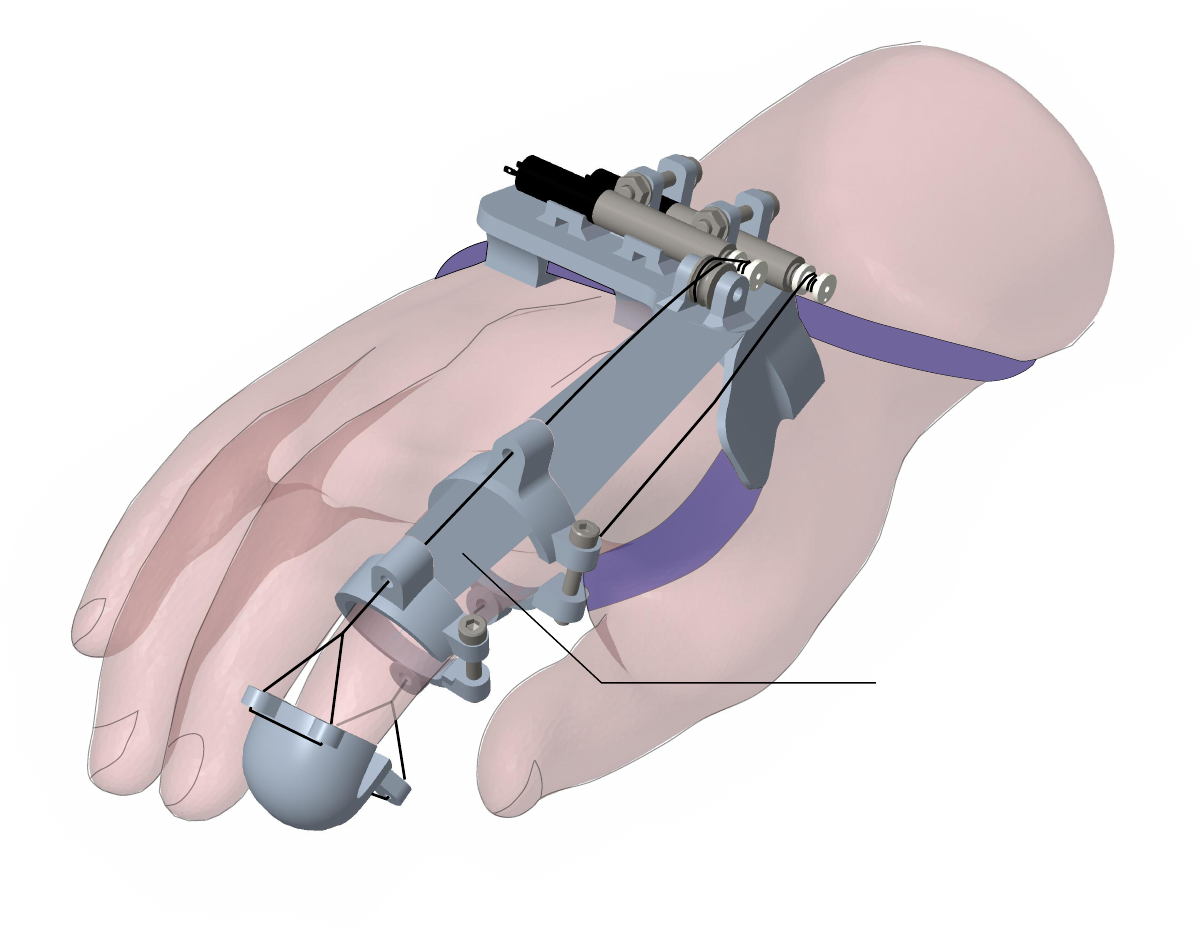}
			\caption{}
			\label{fig:c}
		\end{subfigure}
		\caption{Device design with three different grounding modes: (a) Grounding location is back of the hand, (b) Proximal phalanx is the grounding location, (c) Grounding locations is the Middle phalanx of index finger. In mode (b) and (c), the finger rings are rigidly attached with the base part.}\label{fig:modes}
	\end{figure*} 
	
	We aim to study the effects of different hand-grounding locations on a user's haptic perception by providing kinesthetic feedback on the user's index finger tip. For this purpose, a wearable 2-DoF haptic device is designed that can provide kinesthetic feedback grounded at three different regions of the user's hand (Fig.~\ref{fig:hand}): (i) back of the hand, (ii) proximal phalanx of the index finger, and (iii) middle phalanx of the index finger. The light-weight and modular design provides kinesthetic feedback in two directions: (A) along the index finger axis, and (B) in flexion-extension.
	
	We aim to understand how different hand-grounding locations affect the user's haptic performance and overall experience. To identify the significance and impact of different hand-grounding locations, two psychophysical experiments are carried out using \emph{the method of constant stimuli} \cite{gescheider1985psychophysics} --- one for each feedback direction. The participants were asked, in separate trials, to discriminate the stiffness of two virtual surfaces based on the kinesthetic feedback provided by the hand-grounded device. The Point of Subjective Equality (PSE) and Just Noticeable Difference (JND) were computed to measure the effective sensitivity and precision of the participants' perception of stiffness for each hand-grounding location, in both feedback directions. The PSE gives insight about the accuracy of the applied/perceived feedback, as it represents the point where the comparison stimulus (stiffness) is perceived by the user as identical to the standard stimulus. JND indicates the resolving power of a user and is defined as the minimum change in the stimulus value required to cause a perceptible increase in the sensation~\cite{gescheider1985psychophysics}.
	
	The results show that the choice of grounding location has profound impact on the user's haptic perception (measured through the metrics described above) and experience (based on user ratings). These findings provide important insights for the design of next-generation kinesthetic feedback devices, particularly in terms of grounding of forces, to achieve compelling and natural kinesthetic haptic interaction in real-world haptic and robotic applications. For example, using these findings we can now design hand-grounded wearable kinesthetic devices with appropriate grounding to offer superior haptic performance and user experience. As hand-grounded devices offer comparatively larger operating range and smaller from factor than their world-grounded counterparts, the knowledge related to the choice of hand-grounding location may help to increase the use of wearable kinesthetic devices in the fields of haptics and robot teleoperation. The contribution of this work is the design of a novel wearable kinesthetic device and study results for understanding of the role played by different hand-grounding locations on user stiffness perception.

\section{DEVICE DESIGN \& CONTROL}
	
	\subsection{Design}
\begin{figure}
		\fontfamily{cmss}\selectfont
		\centering
		\def\svgwidth{.8\columnwidth}
		{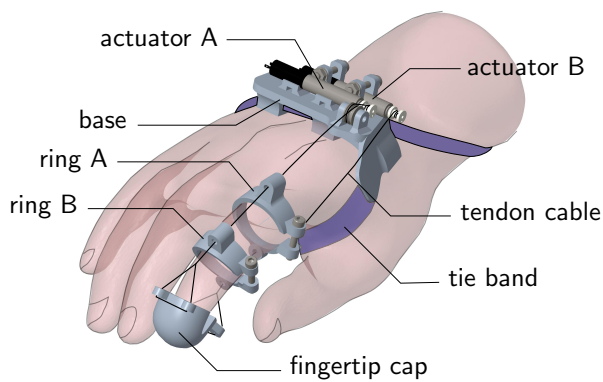}
		\caption{Design: The base is tied against the back of the hand. When tendon cables are pulled/released by actuators A and B, the fingertip cap provides kinesthetic feedback along the finger axis and/or in flexion-extension.}
		\label{fig:device_cad}
	\end{figure}

	The device has a base (Fig.~\ref{fig:device_cad}) that can be tied to the back of the user's hand using a hook-and-loop fastener. 
	It has two rings (A and B) which are fitted to the proximal and middle phalanxes of the index finger. The fingertip cap is connected to actuators A and B through two cables, which route through the passage holes on rings A and B, as shown in Fig.~\ref{fig:device_cad}. When both actuators A and B move in the same direction (clockwise or anti-clockwise), a flexion or extension movement at the finger is produced. When both actuators move in opposite directions, a pull force is generated along the finger axis. 

	To provide hand-grounded kinesthetic feedback at the fingertip, a number of grounding locations can be used. Fig.~\ref{fig:hand} shows the three grounding locations considered in this case: the back of the hand, proximal phalanx of the index finger, and middle phalanx region of the index finger. Another potential location, the palm region, was rejected because such an arrangement may affect the user's ability to open/close the hand and fingers. 
	\begin{figure}
		\fontfamily{cmss}\selectfont
		\centering
		\def\svgwidth{.75\columnwidth}
		{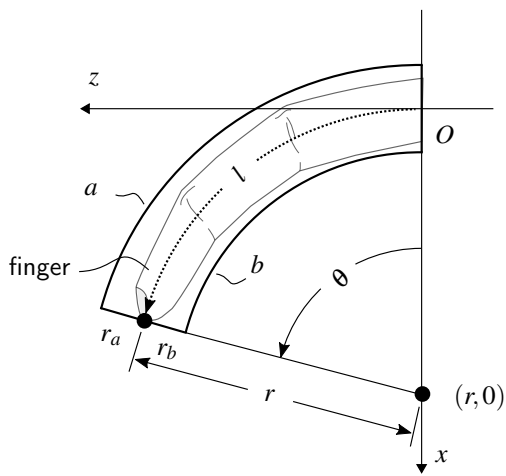}
		\caption{A simplified representation of the device's mechanism as a 2-D piece-wise constant-curvature tendon-driven manipulator. Tendon lengths ($l_a, l_b$), their respective distance from tip center-point ($r_a, r_b$), and the arc parameters: length ($l$) and radius ($r$), are used to determine the tip position and finger configuration in the $x-z$ plane.}
		\label{fig:simp_rep}
	\end{figure}

	To achieve different groundings, the device has three different modes. In mode A (Fig.~\ref{fig:modes}(a)), the back of the hand acts as the grounding location. In mode B (Fig.~\ref{fig:modes}(b)), the base is physically connected to the ring A at the proximal phalanx, providing grounding at this region. In mode C (Fig.~\ref{fig:modes}(c)), the base is rigidly connected with both rings to provide grounding at the middle phalanx region. Different device modes enable execution of different joints of the index finger in the flexion-extension direction. For example, in mode A, the torque is applied at all three joints (MP1, PIP, and DIP). In mode B, only PIP and DIP joints are executed. In mode C, the torque applies only at the DIP joint.
    
\begin{figure}
	\vspace{.2cm}
		\begin{subfigure}{.32\columnwidth}
			\centering
			\includegraphics[width=1\textwidth]{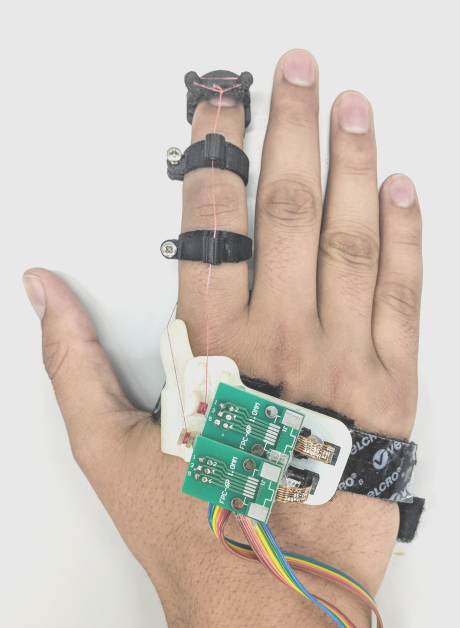}
			\caption{}
			\label{fig:device_a}
		\end{subfigure}
		\begin{subfigure}{.32\columnwidth}
			\centering
			\includegraphics[width=1\textwidth]{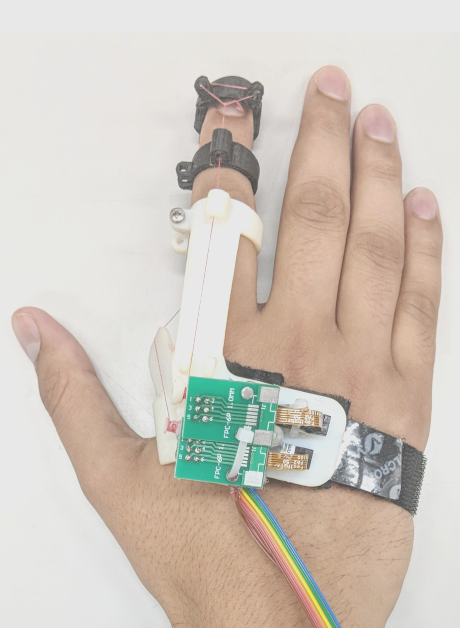}
			\caption{}
			\label{fig:device_b}
		\end{subfigure}
		\begin{subfigure}{.32\columnwidth}
			\centering
			\includegraphics[width=1\textwidth]{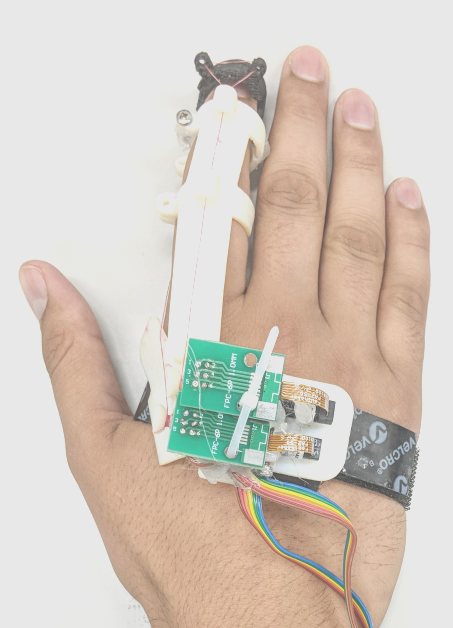}
			\caption{}
			\label{fig:device_c}
		\end{subfigure}
		\caption{Modular versions of the wearable kinesthetic device with grounding locations: (a) Back of the hand, (b) Proximal Phalanx, and (c) Middle Phalanx}
		\label{fig:prototypes}
	\end{figure}

    Based on its kinematic design and actuator specifications, the device can apply, in different modes, a maximum force of 28.9 N along the finger axis and a torque in the range of 80 to 300 N-mm at the fingertip. It is driven by two Faulhaber 0615 4,5S DC-micromotors with 256:1 gearboxes, and 50-counts-per-revolution optical encoders are used for position sensing. The device prototypes with different grounding modes (Fig.~\ref{fig:prototypes}) weigh 31, 43, and 49 grams, respectively. 

	\subsection{Kinematics}
	The device renders forces on the user's index finger by controlling the tendon lengths. To calculate the position and configuration of the finger, we use a robot-independent kinematic mapping between the actuator space and the task space. The obtained homogeneous transformation remains identical for all three grounding modes of our device. It is assumed that the device's tendons, when fit to the user index finger, exhibit a continuum-curve shape. The geometry of this curve allows determination of the tip position and configuration of the finger. Fig.~\ref{fig:simp_rep} shows a simplified representation of the haptic device in such a scheme.
	
	
	\begin{figure*}[ht]
		\vspace{0.2cm}
		\fontfamily{cmss}\selectfont
		\centering
		\def\svgwidth{.9\textwidth}
		{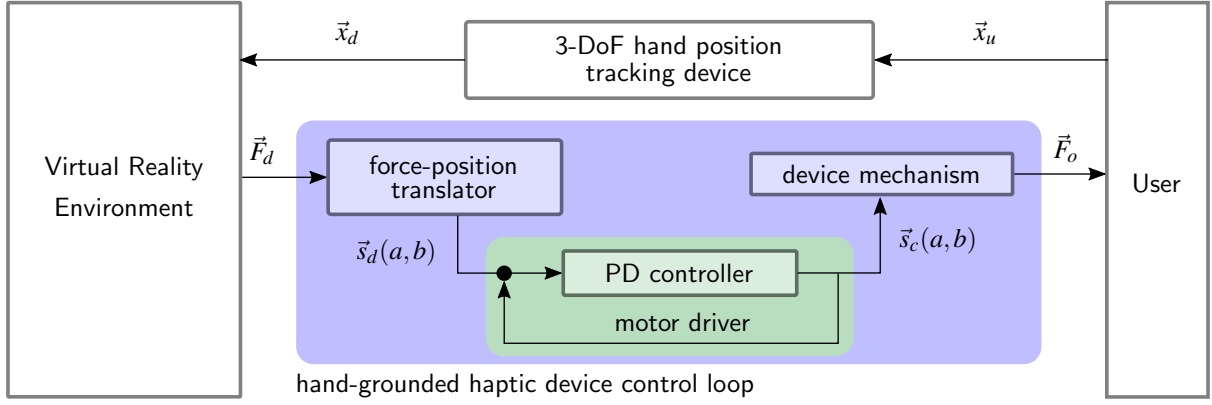}
		\caption{Block diagram of the controller used for rendering force on the user's fingertip. The hand position is tracked by a 3-DoF device, and the interaction forces are calculated as the desired force. Forces applied by the hand-grounded device end-effector on the fingertip through tendon displacements are regulated using a proportional-derivative (PD) controller.}
		\label{fig:control}
	\end{figure*}
	
	As the haptic device aims to provide kinesthetic feedback in two directions (along the finger axis, and in the finger's flexion-extension direction), the kinematic mapping between the inertial frame ($O$) and the fingertip ($p(x,z)$) is described in a 2-D ($x-z$) plane. Tendon lengths ($l_a, l_b$), their respective distance from tip center-point ($r_a, r_b$), and the arc parameters, namely length ($l$) and radius ($r$), are used to determine the tip position and figure configuration in $x-z$ plane. The position of fingertip can be expressed as, 
	\begin{align}
	p(x,z) = \left[r(1 - \cos\theta), r\sin\theta\right]^T.
	\end{align}
	The homogeneous transformation for tendons $a$ and $b$, from $O$ to $p_a(x,z)$ and $p_b(x,z)$ respectively, is 
	
	\begin{align}
		T_j &= \left[\begin{matrix}\operatorname{cos}\left(\theta\right) & 0 & \operatorname{sin}\left(\theta\right) & p_{xj}\\0 & 1 & 0 & 0\\- \operatorname{sin}\left(\theta\right) & 0 & \operatorname{cos}\left(\theta\right) & p_{zj}\\0 & 0 & 0 & 1\end{matrix}\right], \quad (j= a, b),
	\end{align}
	\begin{align}
		p_{xj} &= \left(\frac{l}{\theta} \pm r_{j}\right) \left(1 - \operatorname{cos}\theta\right), \quad (j= a, b), \\
		p_{zj} &= \left(\frac{l}{\theta} \pm r_{j}\right) \operatorname{sin}\theta, \quad (j= a, b).
	\end{align}
	
	The displacements of tendons $a$ and $b$ can be expressed in terms of arc radius and angles as 
	\begin{align}
		s_a = (r + r_a)(\theta_o - \theta_t), \label{eq:s_a} \\
		s_b = (r - r_b)(\theta_o - \theta_t). \label{eq:s_b}
	\end{align}
	where $\theta_o$ is the initial angle angle of tendon $a$ and $\theta_t$ represents the tendon angle at time $t$. 

	\subsection{Control System}
\begin{figure}
		\begin{subfigure}{.49\columnwidth}
			\fontsize{8pt}{6}\selectfont
			\fontfamily{cmss}\selectfont
			\centering
			\def\svgwidth{1\columnwidth}
			{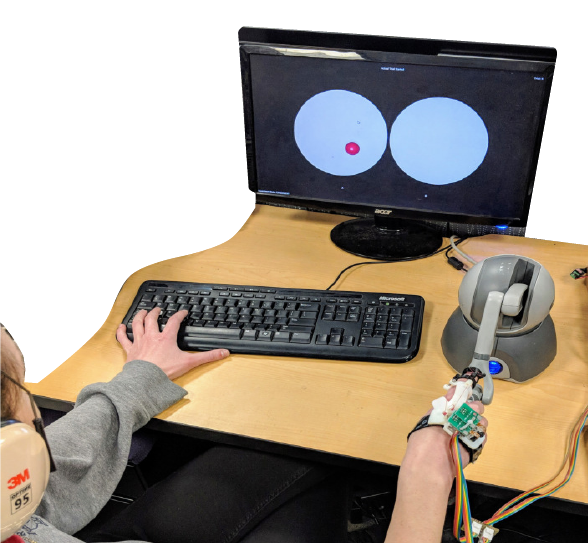}
			\caption{}
			\label{fig:study_b}
		\end{subfigure}
		\begin{subfigure}{.49\columnwidth}
			\fontsize{8pt}{6}\selectfont
			\fontfamily{cmss}\selectfont
			\centering
			\def\svgwidth{1\columnwidth}
			{
\begingroup%
  \makeatletter%
  \providecommand\color[2][]{%
    \errmessage{(Inkscape) Color is used for the text in Inkscape, but the package 'color.sty' is not loaded}%
    \renewcommand\color[2][]{}%
  }%
  \providecommand\transparent[1]{%
    \errmessage{(Inkscape) Transparency is used (non-zero) for the text in Inkscape, but the package 'transparent.sty' is not loaded}%
    \renewcommand\transparent[1]{}%
  }%
  \providecommand\rotatebox[2]{#2}%
  \newcommand*\fsize{\dimexpr\f@size pt\relax}%
  \newcommand*\lineheight[1]{\fontsize{\fsize}{#1\fsize}\selectfont}%
  \ifx\svgwidth\undefined%
    \setlength{\unitlength}{169.44347267bp}%
    \ifx\svgscale\undefined%
      \relax%
    \else%
      \setlength{\unitlength}{\unitlength * \real{\svgscale}}%
    \fi%
  \else%
    \setlength{\unitlength}{\svgwidth}%
  \fi%
  \global\let\svgwidth\undefined%
  \global\let\svgscale\undefined%
  \makeatother%
  \begin{picture}(1,0.91226527)%
    \lineheight{1}%
    \setlength\tabcolsep{0pt}%
    \put(0,0){\includegraphics[width=\unitlength,page=1]{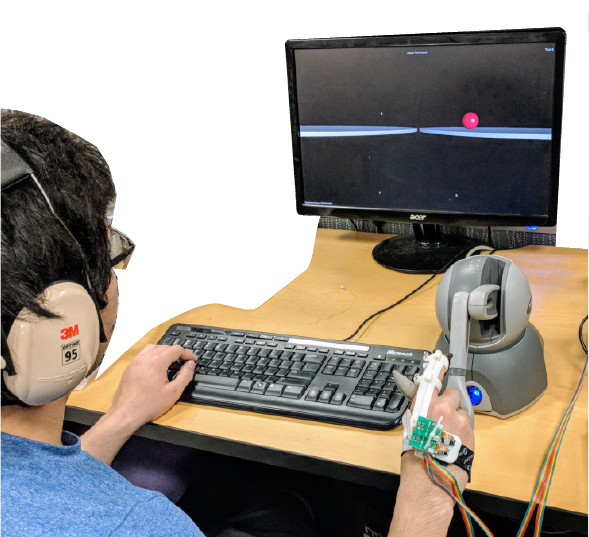}}%
    \put(0.68108066,0.8776621){\color[rgb]{0,0,0}\makebox(0,0)[t]{\lineheight{1.25}\smash{\begin{tabular}[t]{c}virtual surfaces\end{tabular}}}}%
    \put(0.46327347,0.41567671){\color[rgb]{0,0,0}\makebox(0,0)[t]{\lineheight{1.25}\smash{\begin{tabular}[t]{c}keyboard\end{tabular}}}}%
    \put(0,0){\includegraphics[width=\unitlength,page=2]{study_b.pdf}}%
    \put(0.12287586,0.72537548){\color[rgb]{0,0,0}\makebox(0,0)[lt]{\lineheight{1.25}\smash{\begin{tabular}[t]{l}{$z$}\end{tabular}}}}%
    \put(0.24154333,0.67742438){\color[rgb]{0,0,0}\makebox(0,0)[lt]{\lineheight{1.25}\smash{\begin{tabular}[t]{l}{$x$}\end{tabular}}}}%
    \put(0.38318349,0.7936136){\color[rgb]{0,0,0}\makebox(0,0)[lt]{\lineheight{1.25}\smash{\begin{tabular}[t]{l}{$y$}\end{tabular}}}}%
  \end{picture}%
\endgroup%
}
			\caption{}
			\label{fig:study_a}
		\end{subfigure}
		\caption{Experimental setup: A user interacts with the virtual environment through a 3-DoF hand position tracking device (Phantom Omni). The new hand-grounded haptic device provides kinesthetic feedback, and a visual display shows the virtual environment. Participants receive force feedback by touching the two virtual surfaces, one carrying the reference stiffness and the other comparison stiffness in a random order. Participants are required to discriminate the stiffness based on the kinesthetic feedback and record their choice through key presses. (a) Study A (the feedback is rendered along the finger-axis) (b) Study B (the feedback is rendered along flexion-extension movements).}
		\label{fig:studies}
		\vspace{-.3cm}
	\end{figure}

	Using the tendon displacements (\ref{eq:s_a}) and (\ref{eq:s_b}), a separate control is implemented for each of the actuators to apply force and control the user's finger configuration. Fig.~\ref{fig:control} shows the block diagram of the control in the virtual reality setup. The control of the 2-DoF kinesthetic device was achieved by using a Nucleo-F446ZE board by STMicroelectronics\texttrademark connected to a Desktop computer via USB. The microcontroller reads the encoders of the motors and receives the desired force from the virtual environment sent using a PC's serial port. Using this information, it calculates the desired torque output of the motors. The control loop runs at a frequency of approximately 1 kHz. The CHAI3D framework was used to render the 3-D virtual reality environment \cite{conti2005chai3d} using the god-object algorithm \cite{zilles1995god-object} to calculate desired interaction force. The user can move the cursor (red sphere in Fig.~\ref{fig:studies}(a) \& (b)) in 3-D space. Because the wearable device has only 2 DoFs, the third dimension does not give any force feedback to the user. Given the nature of the tasks in the user studies, the third dimension ($y$-axis) is not required to display the force feedback.
	
	
	The user's hand position ($\vec{x_u}$) is tracked using a Phantom Omni haptic device (set up to provide no haptic feedback, just position tracking) from SensAble Technologies, Inc. and sent to the virtual environment as $\vec{x_d}$. The resulting interaction force command from the virtual environment ($\vec{F_d}$) is calculated in the computer and then fed to the hand-grounded haptic device. The device then uses a mapping between the force magnitude and the device tip position (force-position translator) to output the desired tip position to the PD controller which, using the encoders mounted on each motor shaft, can estimate the current tip position and configuration and outputs the appropriate tendon displacements ($\vec{s_c}(a,b)$) to the device's motors. As the tendons shorten, the user's finger tip is moved to the right position, allowing him/her to feel a force.  
	
	The PD controller error and the control law are
	\begin{align}
	e(t) &= y(t) - r(t), \label{eq:error}\\
	U &= K_P e + K_D  \frac{d}{dt} e, \label{eq:law}
	\end{align}
	where, $K_P$ represents the proportional gain and $K_D$ is the derivative gain. $e(t)$ is the position error, $y(t)$ represents the motor shaft position, and $r(t)$ is the reference position calculated from the desired tendon displacements ($\vec{s_d}(a,b)$ in Fig.~\ref{fig:control}).

\section{USER STUDY}
	To evaluate the effects of the three different hand-grounding locations on the user's haptic perception and experience, we conducted two separate user studies (Study A \& Study B); one for each haptic feedback DoF provided by the hand-grounded device. In Study A, the kinesthetic feedback is provided along the axis of the user's index finger. In Study B, the feedback is provided along the flexion-extension movement of the finger. The purpose of evaluating each feedback DoF separately is to develop a clear understanding of the relation between the hand-grounding location and the corresponding feedback direction. 
	
	\subsection{Study A: Feedback Along the Finger Axis}
	\subsubsection{Experimental Setup}
	13 subjects (9 males and 4 females) participated in this study, which was approved by the Stanford University Institutional Review Board. The  metrics were PSE and JND of stiffness perception while the hand-grounded device was set up for each of the three grounding locations (back of the hand, proximal, and middle phalanx of the index finger). All subjects participated in the experiment after giving informed consent, under a protocol approved by the Stanford University Institutional Review Board. The participants used the hand-grounded haptic device on their right hand and performed tasks in a virtual environment, while holding the stylus of the Phantom Omni device in the same hand (Fig.~\ref{fig:studies}(a)). A pilot study was conducted to determine a convenient posture to hold the Phantom Omni stylus while the kinesthetic device is donned to the index finger. In the user studies, the participants were instructed to hold the Phantom Omni device in that predefined way to make sure that its stylus does not come into contact with the wearable kinesthetic device. 
	
	\subsubsection{Experimental Procedure}
	Each participant used the haptic device configured for each of the three hand-grounding locations in a predetermined order to minimize the effect of selection bias. As mentioned earlier, a Phantom Omni device was used to track the user hand position during the experiments as shown in Fig.~\ref{fig:studies}(a). The Phantom Omni only determined the user hand position, while the kinesthetic feedback was rendered by the hand-grounded haptic device. 
	Participants were wore ear protection to suppress the motor noise in order to avoid sound cues. After the experiments were completed, the participants rated the realism of haptic feedback and comfort/ease-of-use for all three devices with different hand-grounding locations on a scale of 1-7: 1 meaning `not real' and 7 meaning `real', or 1 for `not comfortable' and 7 for `comfortable.' The realism was with respect to the users' feeling as if they would be pressing against a very smooth real surface using their right-hand’s index finger.
	
	\subsubsection{Method}
	We conducted a \emph{two-alternative forced-choice} experiment following the \emph{method of constant stimuli}~\cite{gescheider1985psychophysics}. Subjects were asked to freely explore and press against the two virtual surfaces shown on  the virtual environment display and state which surface felt stiffer. In each trial, one surface presented a reference stiffness value while the other presented a comparison stiffness value. The reference stiffness value was selected to be 100.0 N/m.
The reference value was included as one comparison value, and the other comparison values were then chosen to be equally spaced: 10, 28, 46, 64, 82, 100, 118, 136, 154, 172, and 190 N/m.
%

	Each of the eleven comparison values was presented ten times in random order for each of the three hand-grounded haptic devices over the course of one study. Each participant completed a total of 110 trials for each grounding mode (330 trials for the entire study). The participants used the kinesthetic feedback from the hand-grounded device to explore the virtual surfaces until a decision was made; they recorded their responses by pressing designated keyboard keys, corresponding to which virtual surface they thought felt stiffer. Subject responses and force/torque data were recorded after every trial. There was no time limit for each trial, and participants were asked to make their best guess if the decision seemed too difficult. Subjects were given an optional two-minute break after every fifty-five trials, and a ten-minute break after the completion of each grounding mode.
			
	\subsection{Study B: Feedback in the Flexion-Extension Direction}
	In study B, the kinesthetic feedback was rendered along the flexion extension movement direction of the index finger. A total of 14 subjects (9 males and 5 females) participated, and the study was approved by the Stanford University Institutional Review Board. The procedure was the same as in Study A. However, in Study B the virtual surfaces were presented lying in the horizontal plane (Fig.~\ref{fig:studies}(b)) to make the haptic feedback intuitive for the user.

	\begin{figure*}
		\renewcommand\thesubfigure{\roman{subfigure}}
		\begin{subfigure}{.32\textwidth}
			\fontfamily{cmss}\selectfont
			\centering
			\includegraphics[width=1\columnwidth]{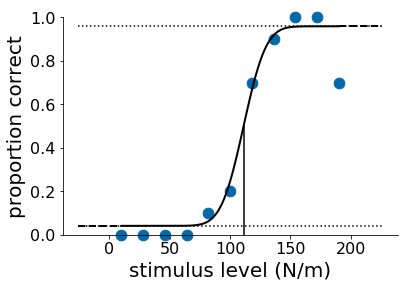}
			\caption{Back of the hand}
			\label{fig:curve_a}
		\end{subfigure}
		\begin{subfigure}{.32\textwidth}
			\fontfamily{cmss}\selectfont
			\centering
			\includegraphics[width=1\columnwidth]{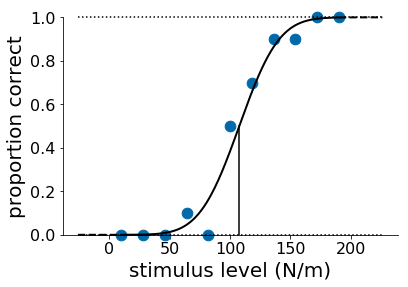}
			\caption{Proximal Phalanx}
			\label{fig:curve_b}
		\end{subfigure}
		\begin{subfigure}{.32\textwidth}
			\fontfamily{cmss}\selectfont
			\centering
			\includegraphics[width=1\columnwidth]{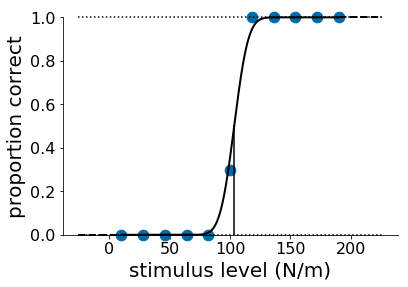}
			\caption{Middle Phalanx}
			\label{fig:curve_c}
		\end{subfigure}
		\caption{Example psychophysical data and psychometric function fits for a representative subject in Study A, with grounding locations: (i) back of the hand, (ii) proximal phalanx, and (iii) middle phalanx of the index finger. Each data point represents the 'yes' proportion of the user responses over 10 trials. The user identified the difference between the reference and comparison stimulus values correctly 90 \% of the time for grounding location (i), 94 \% of the time for location (ii), and 98 \% of the times for location (iii).}
		\label{fig:curves}
	\end{figure*}
	\begin{table*}
		\vspace{.3cm}
		\caption{Results of the two psychophysical experiments for stiffness discrimination. In Study A, the hand-grounded haptic device provided feedback along the axis of the finger with three different grounding locations. In Study B, the feedback direction was flexion extension movement of the index finger.}
		\begin{tabularx}{\linewidth}{|c|r|XX|XX|XX|}
			\hline 
			\multicolumn{2}{|c|}{Grounding Location} & \multicolumn{2}{c}{Back of the Hand} & \multicolumn{2}{c}{Proximal Phalanx} & \multicolumn{2}{c|}{Mid Phalanx} \\
			\hline 
			\hline 
			& Subject No.&   PSE (N/m) &   JND (N/m) &   PSE (N/m) &   JND (N/m) &   PSE (N/m) &   JND (N/m) \\
			\multirow{14}{*}{\rotatebox[origin=c]{90}{\parbox[c]{1cm}{\centering Study A}}} &  1 & 154.42 & 57.15 & 150.74 & 47.35 & 120.34 & 41.32 \\
			&  2 & 111.17 &  9.86 & 107.58 & 17.44 & 103.37 &  6.2  \\
			&  3 & 139.75 & 48.5  & 129.95 & 34.27 &  81.3  &  7.32 \\
			&  4 &  87.56 &  6.07 &  92.95 &  5.72 & 102.27 & 20.24 \\
			&  5 &  98.62 & 32.37 &  85.2  &  6.79 & 114.85 & 19.04 \\
			&  6 & 113.23 & 13.08 & 104.68 & 25.28 & 100.74 & 20.1  \\
			&  7 &  99.6  & 13.21 & 108.99 &  7.97 &  94.68 &  9.48 \\
			&  8 & 118.75 & 46.88 & 128    & 31.13 & 109.11 & 38.95 \\
			&  9 & 115.5  & 20.23 & 103.4  & 10.16 & 105.51 & 19.72 \\
			& 10 & 114.64 & 12.93 & 103.46 & 12.15 & 116.09 & 19    \\
			& 11 &  85.87 &  7.32 & 108.22 & 12.65 & 117.25 & 17.41 \\
			& 12 &  92.94 &  6.66 & 114.14 & 22.17 &  91.54 & 15.65 \\
			\cline{2-8}
			& Mean      & 111.004 & 22.855 & 111.442 & 19.423 & 104.754 & 19.536 \\
			& Std. Dev. &  19.581 & 17.677 &  16.843 & 12.404 &  11.178 & 10.411 \\
			\hline
			\hline
			\multirow{15}{*}{\rotatebox[origin=c]{90}{\parbox[c]{1cm}{\centering Study B}}} &  1 &  96.06 &  0.17 & 101.2  & 12.26 & 103.8  & 13.75 \\
			&  2 &  92.71 & 13.51 & 104.59 & 13.02 &  92.79 &  8.64 \\
			&  3 & 125.72 & 40.06 &  94.9  & 28.36 & 121.88 & 45.98 \\
			&  4 & 104.87 &  6.41 &  98.32 & 14.36 &  87.34 &  6.75 \\
			&  5 & 115.52 & 41.74 & 114.25 & 40.84 & 102.2  & 20    \\
			&  6 & 122.16 & 33.52 &  90.2  & 16.44 & 119.91 & 16.88 \\
			&  7 & 121.54 & 20.98 & 116.83 & 36.64 & 125.8  & 64.77 \\
			&  8 & 105.76 & 10.69 & 108.89 & 11.67 & 104.6  & 26.4  \\
			&  9 &  98.85 & 25.9  & 100    & 16.83 & 117.3  & 25.28 \\
			& 10 &  99.45 & 41.48 &  95.07 & 22.3  & 104.98 & 24.18 \\
			& 11 & 138.96 & 41.62 & 127    & 30.4  & 122.64 & 24.03 \\
			& 12 & 113.34 & 38.78 &  95.35 & 45.78 & 120.79 & 60.43 \\
			\cline{2-8}
			& Mean      & 111.245 & 26.238 & 103.883 & 24.075 & 110.336 & 28.091 \\
			& Std. Dev. &  13.426 & 14.761 &  10.435 & 11.520 &  12.181 & 18.222 \\
			\hline
		\end{tabularx}
	\end{table*}

\section{RESULTS \& DISCUSSION}
	For both user studies, we determined the number of times each participant responded that the comparison value of stiffness was greater than the reference stiffness value. A psychometric function was then fit for each participant's response data to plot a psychometric curve, using the python-psignifit 4 library (https://github.com/wichmann-lab/python-psignifit). Data from twenty-four out of the twenty-seven subjects fit sufficiently to psychometric functions and the mean JNDs and PSEs for both experiments were determined. Example plots for a representative subject are shown in Fig.~\ref{fig:curves}. Three relevant values: the PSE, the stimulus value corresponding to a proportion of 0.25 ($J_{25}$), and the stimulus value corresponding to a proportion of 0.75 ($J_{75}$) were determined. The JND is defined as the mean of the differences between the PSE and the two J values $J_{25}$ and $J_{75}$: 

	\begin{align}
		JND = \frac{(PSE - J_{25}) + (J_{75} - PSE)}{2}.
	\end{align}
	
%
	\begin{figure}
		\centering
		\includegraphics[width=1\columnwidth]{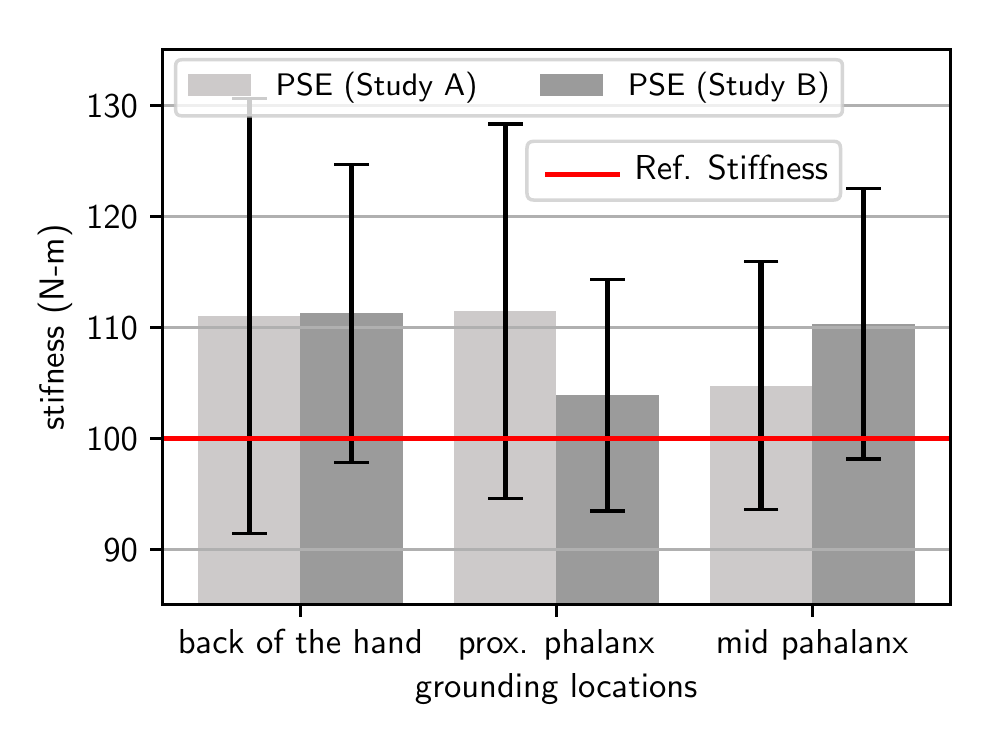}
		\caption{Point of Subjective Equality (PSE) for both feedback DoFs (Study A and B) against each of the three considered hand-grounding locations. Error bars indicate the standard deviation.}
		\label{fig:pse}
	\end{figure}

    \begin{figure}
    	\vspace{-.3cm}
		\centering
		\includegraphics[width=1\columnwidth]{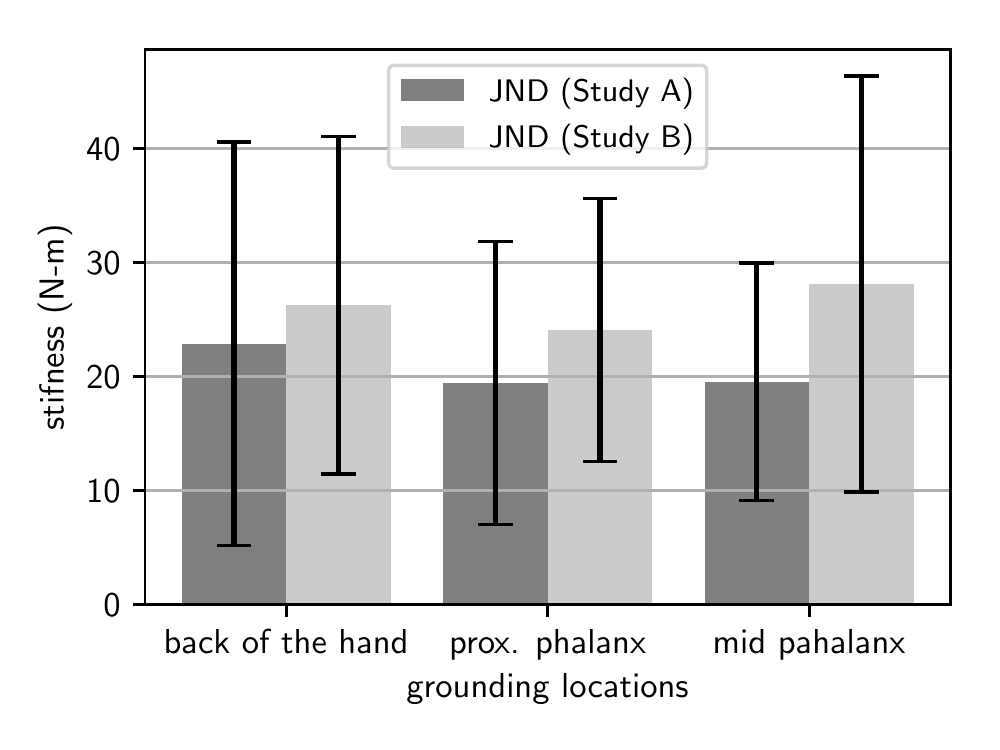}
		\caption{Just Noticeable Differences (JNDs) for both feedback DoFs (Study A and B) against each considered hand-grounding locations. 
		}
		\label{fig:jnd}
		\vspace{-.3cm}
	\end{figure}   
	
	The PSE and JND results of the psychophysical experiments for both studies are summarized in Table 1. Because these studies use a single reference force, the Weber Fractions (WFs) are simply the JNDs scaled by the reference value. Therefore, we do not report WF separately.
	
\begin{figure}
		\centering
		\vspace{-0.65cm}
		\includegraphics[width=1\columnwidth]{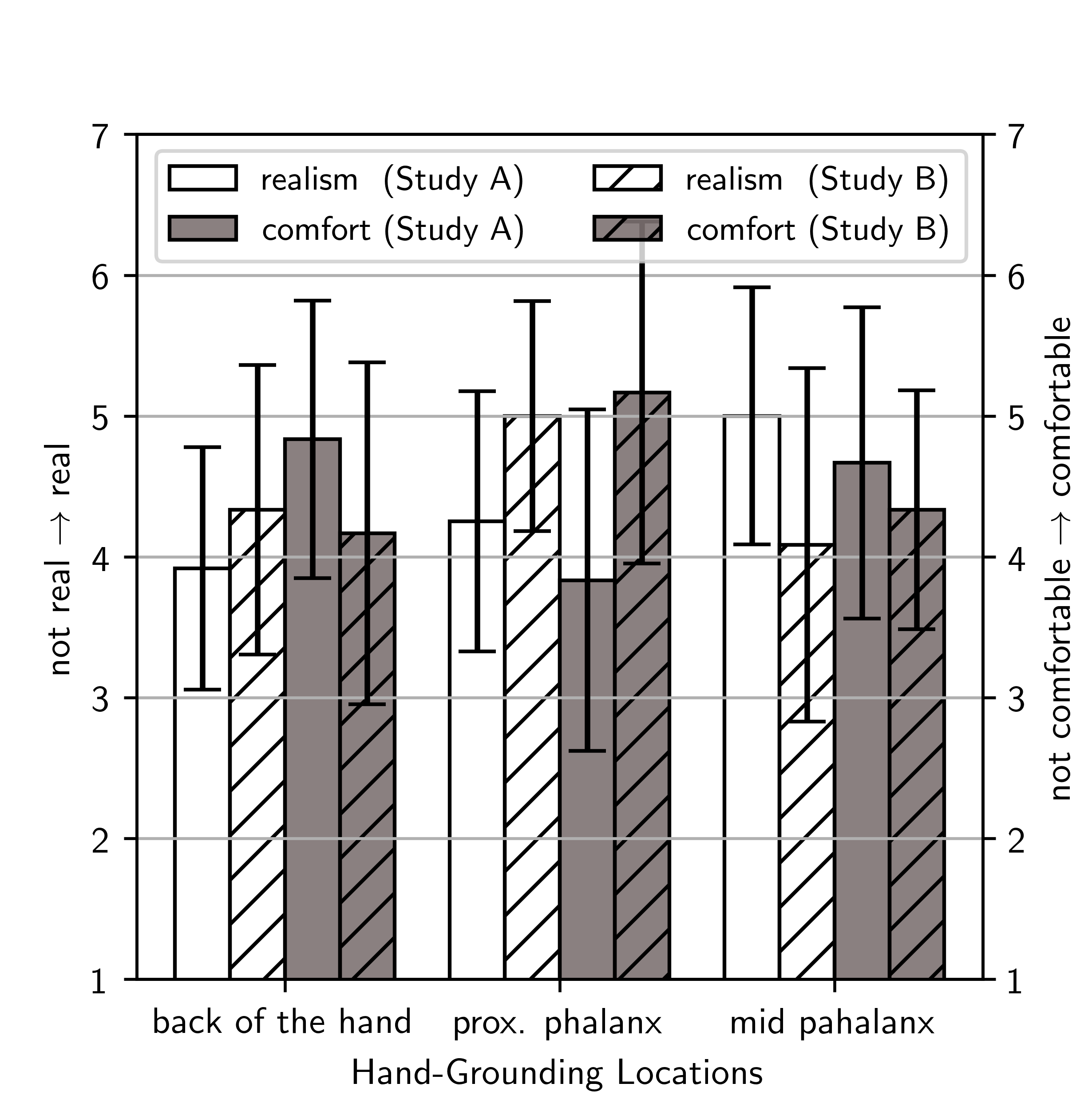}
		\caption{Mean user ratings for the realism of feedback and comfort/ease-of-use against each of the three hand-grounding locations. Error bars indicate standard deviations.}
		\label{fig:ratings}
	\end{figure}

	In Study A, the best average PSE (closer to the reference value) for stiffness perception among all three grounding locations is found for the middle phalanx location of the index finger (104.75 N/m), shown in Fig.~\ref{fig:pse}. This indicates that the grounding location closer to the fingertip helps users to perceive the stiffness more accurately. This is also supported by the user ratings for the realism of kinesthetic feedback, as shown in Fig.~\ref{fig:ratings}. The smallest average JND was found for grounding at proximal phalanx (19.42 N/m), which is closely followed by average JND values for grounding location at the proximal phalanx (19.54 N/m). Like the PSE, the average JND showed largest value for back of the hand grounding location (see Fig.~\ref{fig:jnd}). This indicates that the proximal and middle phalanx are preferable locations, in the given order, to have a more realistic and accurate feedback perception. However, the user ratings for the comfort and ease-of-use indicate that the back of the hand is a more desirable grounding location.
	
	In Study B, the best average PSE (closer to the reference value) for stiffness perception among all three grounding locations is found in the grounding at the proximal phalanx of the index finger (103.88 N/m), shown in Fig.~\ref{fig:pse}. This grounding location also results in the smallest average JND value (24.07 N/m) among all three grounding locations. The user ratings for kinesthetic feedback realism and the comfort/ease of use, as show in Fig.~\ref{fig:ratings}, also rate this location as the best to impart most realistic and comfortable haptic experience. The second best location in terms of average JND value is the back of the hand. This holds for the feedback realism ratings as well. The grounding location with least realistic feedback ratings and largest average JND (28.09 N/m) was the proximal phalanx location.
	
	If we compare the average JND values across both studies, the values for feedback along the finger-axis (Study A) are significantly smaller than that of the feedback along flexion-extension direction (study B). This indicates that the haptic device was able to provide better haptic feedback in case of Study A, i.e. along the axis of the index-finger. The reason for this probably relates to the simpler nature of this feedback direction where the finger configuration remains unchanged during all modes. However, the realism and comfort ratings show a distinct pattern; realism is higher for Study B (kinesthetic feedback along flexion-extension) when the grounding locations are the back of the hand and proximal phalanx. The realism in case of Study A is higher than that of the B when grounding location is middle phalanx. This again depends on the different nature of the second feedback DoF, where the finger configuration has to change in order to render a torque at the finer joints. The use of the Phantom Omni for tracking may introduce some passive forces that introduce variance in the study. Despite this, we observed significant performance differences among the studied grounding locations.
	
	On the other hand, the comfort/ease-of-use ratings are higher for Study A than B, when the grounding locations are the back of the hand and the middle phalanx, respectively. The feedback in flexion-extension movement (Study B) has shown higher comfort ratings than for the finger-axis direction (Study A) when grounding is set as the proximal phalanx region. The highest comfort rating among all grounding locations across both studies is given to the proximal phalanx, and that is for the feedback along the flexion-extension movement. Similarly, the highest comfort rating is given to the same grounding location, i.e., proximal phalanx, across both studies, and that too is for the feedback along flexion-extension direction. 
	
%


\section{CONCLUSION}
	A novel hand-grounded kinesthetic feedback device was created for studying the effect of different grounding locations on the user's haptic experience. The device can provide kinesthetic feedback along the user's index finger, and in its flexion-extension movement direction. Two psychophysical experiments -- one for each feedback DoF -- were conducted to evaluate the user's haptic performance and experience. It is shown that the choice of grounding-location in wearable haptic devices has significant impact over the user haptic perception of stiffness. The realism of the haptic feedback increases, while the comfort level decrease, as the grounding location moves closer to the fingertip. The relationship between the grounding-location and user haptic perception is similar in both feedback directions. If the design objective is to achieve maximum comfort, feedback realism, and best haptic perception in both DoFs simultaneously, it is recommended to have grounding at the proximal phalanx region of the finger. 
	
	These findings about the choice and impact of different hand-grounding locations give important insights for designing next-generation wearable kinesthetic devices, and to have better performance in a wide range of applications, such as virtual reality and robot teleoperation.  
	In the future, we plan to conduct further experiments to explore the effects of these hand-grounding locations when the kinesthetic feedback is applied to both DoFs simultaneously. 









\bibliographystyle{IEEEtran}
\bibliography{handgrounded_sajid}

\begin{thebibliography}{10}
\providecommand{\url}[1]{#1}
\csname url@samestyle\endcsname
\providecommand{\newblock}{\relax}
\providecommand{\bibinfo}[2]{#2}
\providecommand{\BIBentrySTDinterwordspacing}{\spaceskip=0pt\relax}
\providecommand{\BIBentryALTinterwordstretchfactor}{4}
\providecommand{\BIBentryALTinterwordspacing}{\spaceskip=\fontdimen2\font plus
\BIBentryALTinterwordstretchfactor\fontdimen3\font minus
  \fontdimen4\font\relax}
\providecommand{\BIBforeignlanguage}[2]{{%
\expandafter\ifx\csname l@#1\endcsname\relax
\typeout{** WARNING: IEEEtran.bst: No hyphenation pattern has been}%
\typeout{** loaded for the language `#1'. Using the pattern for}%
\typeout{** the default language instead.}%
\else
\language=\csname l@#1\endcsname
\fi
#2}}
\providecommand{\BIBdecl}{\relax}
\BIBdecl

\bibitem{pacchierotti2017wearable}
C.~Pacchierotti, S.~Sinclair, M.~Solazzi, A.~Frisoli, V.~Hayward, and
  D.~Prattichizzo, ``Wearable haptic systems for the fingertip and the hand:
  Taxonomy, review, and perspectives,'' \emph{IEEE Transactions on Haptics},
  vol.~10, no.~4, pp. 580--600, 2017.

\bibitem{sucho@2016}
J.~M. Suchoski, A.~Barron, C.~Wu, Z.~F. Quek, S.~Keller, and A.~M. Okamura,
  ``Comparison of kinesthetic and skin deformation feedback for mass
  rendering,'' in \emph{IEEE International Conference on Robotics and
  Automation}, 2016, pp. 4030--4035.

\bibitem{okamura2009haptic}
A.~M. Okamura, ``Haptic feedback in robot-assisted minimally invasive
  surgery,'' \emph{Current Opinion in Urology}, vol.~19, no.~1, p. 102, 2009.

\bibitem{burdea1999keynote}
G.~C. Burdea, ``Keynote address: haptics feedback for virtual reality,'' in
  \emph{Proceedings of International Workshop on Virtual Prototyping}, 1999,
  pp. 87--96.

\bibitem{jadhav2017soft}
S.~Jadhav, V.~Kannanda, B.~Kang, M.~T. Tolley, and J.~P. Schulze, ``Soft
  robotic glove for kinesthetic haptic feedback in virtual reality
  environments,'' \emph{Electronic Imaging}, vol. 2017, no.~3, pp. 19--24,
  2017.

\bibitem{fontana2009mechanical}
M.~Fontana, A.~Dettori, F.~Salsedo, and M.~Bergamasco, ``Mechanical design of a
  novel hand exoskeleton for accurate force displaying,'' in \emph{IEEE
  International Conference on Robotics and Automation}, 2009, pp. 1704--1709.

\bibitem{springer2002design}
S.~L. Springer and N.~J. Ferrier, ``Design and control of a force-reflecting
  haptic interface for teleoperational grasping,'' \emph{Journal of Mechanical
  Design}, vol. 124, no.~2, pp. 277--283, 2002.

\bibitem{leonardis2015emg}
D.~Leonardis, M.~Barsotti, C.~Loconsole, M.~Solazzi, M.~Troncossi, C.~Mazzotti,
  V.~P. Castelli, C.~Procopio, G.~Lamola, C.~Chisari \emph{et~al.}, ``An
  emg-controlled robotic hand exoskeleton for bilateral rehabilitation,''
  \emph{IEEE Transactions on Haptics}, vol.~8, no.~2, pp. 140--151, 2015.

\bibitem{nycz2016design}
C.~J. Nycz, T.~B{\"u}tzer, O.~Lambercy, J.~Arata, G.~S. Fischer, and
  R.~Gassert, ``Design and characterization of a lightweight and fully portable
  remote actuation system for use with a hand exoskeleton,'' \emph{IEEE
  Robotics and Automation Letters}, vol.~1, no.~2, pp. 976--983, 2016.

\bibitem{allotta2015development}
B.~Allotta, R.~Conti, L.~Governi, E.~Meli, A.~Ridolfi, and Y.~Volpe,
  ``Development and experimental testing of a portable hand exoskeleton,'' in
  \emph{IEEE/RSJ International Conference on Intelligent Robots and Systems},
  2015, pp. 5339--5344.

\bibitem{ma2015rml}
Z.~MA and P.~Ben-Tzvi, ``{RML Glove}---{An} exoskeleton glove mechanism with
  haptics feedback,'' \emph{IEEE/ASME Transactions on Mechatronics}, vol.~20,
  no.~2, pp. 641--652, 2015.

\bibitem{kim2016hapthimble}
H.~Kim, M.~Kim, and W.~Lee, ``Hapthimble: A wearable haptic device towards
  usable virtual touch screen,'' in \emph{Proc. CHI Conference on Human Factors
  in Computing Systems}.\hskip 1em plus 0.5em minus 0.4em\relax ACM, 2016, pp.
  3694--3705.

\bibitem{fu2011design}
Y.~Fu, Q.~Zhang, F.~Zhang, and Z.~Gan, ``Design and development of a hand
  rehabilitation robot for patient-cooperative therapy following stroke,'' in
  \emph{IEEE International Conference on Mechatronics and Automation}, 2011,
  pp. 112--117.

\bibitem{lambercy2013design}
\BIBentryALTinterwordspacing
O.~Lambercy, D.~Schr\"{o}der, S.~Zwicker, and R.~Gassert, ``Design of a thumb
  exoskeleton for hand rehabilitation,'' in \emph{Proc. 7th International
  Convention on Rehabilitation Engineering and Assistive Technology}, ser.
  i-CREATe '13.\hskip 1em plus 0.5em minus 0.4em\relax Kaki Bukit TechPark II,,
  Singapore: Singapore Therapeutic, Assistive \& Rehabilitative Technologies
  (START) Centre, 2013, pp. 41:1--41:4. [Online]. Available:
  \url{http://dl.acm.org/citation.cfm?id=2567429.2567477}
\BIBentrySTDinterwordspacing

\bibitem{stergiopoulos2003design}
P.~Stergiopoulos, P.~Fuchs, and C.~Laurgeau, ``Design of a 2-finger hand
  skeleton for {VR} grasping simulation,'' in \emph{Eurohaptics}, 2003, pp.
  80--93.

\bibitem{lelieveld2006design}
M.~J. Lelieveld, T.~Maeno, and T.~Tomiyama, ``Design and development of two
  concepts for a 4 {DOF} portable haptic interface with active and passive
  multi-point force feedback for the index finger,'' in \emph{Proc.
  International Design Engineering Technical Conferences and Computers and
  Information in Engineering Conference}.\hskip 1em plus 0.5em minus
  0.4em\relax ASME, 2006, pp. 547--556.

\bibitem{cempini2015powered}
M.~Cempini, M.~Cortese, and N.~Vitiello, ``A powered finger--thumb wearable
  hand exoskeleton with self-aligning joint axes,'' \emph{IEEE/ASME
  Transactions on Mechatronics}, vol.~20, no.~2, pp. 705--716, 2015.

\bibitem{agarwal2015index}
P.~Agarwal, J.~Fox, Y.~Yun, M.~K. O’Malley, and A.~D. Deshpande, ``An index
  finger exoskeleton with series elastic actuation for rehabilitation: Design,
  control and performance characterization,'' \emph{The International Journal
  of Robotics Research}, vol.~34, no.~14, pp. 1747--1772, 2015.

\bibitem{aiple2013pushing}
M.~Aiple and A.~Schiele, ``Pushing the limits of the
  {CyberGrasp}\texttrademark~for haptic rendering,'' in \emph{IEEE
  International Conference on Robotics and Automation}, 2013, pp. 3541--3546.

\bibitem{tanaka2002wearable}
Y.~Tanaka, H.~Yamauchi, and K.~Amemiya, ``Wearable haptic display for immersive
  virtual environment,'' in \emph{Proc. International Symposium on Fluid
  Power}, no. 5-2.\hskip 1em plus 0.5em minus 0.4em\relax The Japan Fluid Power
  System Society, 2002, pp. 309--314.

\bibitem{polygerinos2015soft}
P.~Polygerinos, Z.~Wang, K.~C. Galloway, R.~J. Wood, and C.~J. Walsh, ``Soft
  robotic glove for combined assistance and at-home rehabilitation,''
  \emph{Robotics and Autonomous Systems}, vol.~73, pp. 135--143, 2015.

\bibitem{stetten2011hand}
G.~Stetten, B.~Wu, R.~Klatzky, J.~Galeotti, M.~Siegel, R.~Lee, F.~Mah,
  A.~Eller, J.~Schuman, and R.~Hollis, ``Hand-held force magnifier for surgical
  instruments,'' \emph{Information Processing in Computer-Assisted
  Interventions}, pp. 90--100, 2011.

\bibitem{bouzit2002rutgers}
M.~Bouzit, G.~Burdea, G.~Popescu, and R.~Boian, ``The {Rutgers Master II}-new
  design force-feedback glove,'' \emph{IEEE/ASME Transactions on Mechatronics},
  vol.~7, no.~2, pp. 256--263, 2002.

\bibitem{choi2018claw}
I.~Choi, E.~Ofek, H.~Benko, M.~Sinclair, and C.~Holz, ``Claw: A multifunctional
  handheld haptic controller for grasping, touching, and triggering in virtual
  reality,'' in \emph{Proc. CHI Conference on Human Factors in Computing
  Systems}, 2018, pp. 654:1--654:13.

\bibitem{choi2016wolverine}
I.~Choi, E.~W. Hawkes, D.~L. Christensen, C.~J. Ploch, and S.~Follmer,
  ``Wolverine: A wearable haptic interface for grasping in virtual reality,''
  in \emph{IEEE/RSJ International Conference on Intelligent Robots and
  Systems}, 2016, pp. 986--993.

\bibitem{khurshid2014wearable}
R.~M. Pierce, E.~A. Fedalei, and K.~J. Kuchenbecker, ``A wearable device for
  controlling a robot gripper with fingertip contact, pressure, vibrotactile,
  and grip force feedback,'' in \emph{IEEE Haptics Symposium}, 2014, pp.
  19--25.

\bibitem{gescheider1985psychophysics}
G.~Gescheider, \emph{Psychophysics: method, theory, and application},
  2nd~ed.\hskip 1em plus 0.5em minus 0.4em\relax L. Erlbaum Associates, 1985.

\bibitem{conti2005chai3d}
D.~M. F.~Conti, F.~Barbagli and C.~Sewell, ``{CHAI 3D} - an open-source library
  for the rapid development of haptic scenes,'' in \emph{IEEE World Haptics},
  2005, pp. 21--29.

\bibitem{zilles1995god-object}
C.~B. Zilles and J.~K. Salisbury, ``A constraint-based god-object method for
  haptic display,'' in \emph{IEEE/RSJ International Conference on Intelligent
  Robots and Systems}, vol.~3, 1995, pp. 146--151 vol.3.

\end{thebibliography}

\end{document}